\documentclass[11pt]{article}
\usepackage[a4paper,textwidth=126mm,textheight=195mm,centering]{geometry}
\usepackage{amsmath,amssymb,booktabs,graphicx,microtype,float,array}
\usepackage[hidelinks]{hyperref}
\usepackage[font=small,labelfont=bf]{caption}
\title{When does a spectral prior help graph learning? Connectivity-loss estimation under road-network disruptions}
\author{Van-Truong Le$^{*}$\\
\small Faculty of Information Technology, University of Science,\\
\small Viet Nam National University Ho Chi Minh City, Viet Nam\\
\small $^{*}$Corresponding author:
\href{mailto:23120181@student.hcmus.edu.vn}{23120181@student.hcmus.edu.vn}\\
\small E-mail: \href{mailto:lvtruong@selab.hcmus.edu.vn}{lvtruong@selab.hcmus.edu.vn}\\
\small ORCID: \href{https://orcid.org/0009-0008-7015-7392}{0009-0008-7015-7392}}
\date{}

\begin{document}
\maketitle

\begin{abstract}
Rapid evaluation of many simultaneous road-link disruptions requires a useful
compromise between exact spectral recomputation and local approximation. We
estimated the relative loss of algebraic connectivity after multi-edge deletion
using graph neural networks (GNNs) that learn a bounded correction to a
first-order Fiedler sensitivity. The design was tested under independent,
spatially clustered, and edge-betweenness-targeted failures, with graph-disjoint
synthetic splits and zero-shot transfer to 13 OpenStreetMap (OSM) areas in six
countries (48--1,259 nodes). GCN, GraphSAGE, and explicitly edge-aware MPNN
backbones and analytical baselines isolated the residual prior. For expanded OSM
analyses, uncertainty used an area-clustered hierarchical bootstrap, with seeds
nested within area. Across the expanded
OSM panel, residual GCN improved spatial-failure MAE by 0.0391 (95\% hierarchical
interval 0.0151--0.0662), while residual GraphSAGE improved targeted-failure MAE
by 0.0257 (0.0095--0.0446). Edge-MPNN residual differences were positive but
imprecise. Second-order perturbation improved first-order MAE by only
0.0028--0.0053. Correction slopes fell from 0.84--0.95 under independent or
spatial transfer to 0.51--0.58 under targeted transfer for GCN/GraphSAGE,
directly quantifying residual shrinkage around systematic prior error.
The matched per-regime leave-one-country-out OSM-to-OSM transfer was mixed: residual
GCN improved targeted-failure MAE by 0.0622 (0.0169--0.1153), but worsened the
spatial point estimate; the same sign pattern appears under joint training.
Controlled sparse eigensolver scaling extended to 20,000 nodes and separated the one-time
spectral setup from the amortized cost. Within the limited geographic
clusters, these results characterize the spectral residual as a potentially useful
but domain-sensitive inductive bias and delimit
its applicability to structural connectivity rather than hazard or traffic-flow
prediction. Code, cached networks, and reproducibility artefacts are permanently
archived at \href{https://doi.org/10.5281/zenodo.22307723}{doi:10.5281/zenodo.22307723}.
The development repository is available on
\href{https://github.com/lee-vtruong/ResiliRoad}{GitHub}.
\end{abstract}


\section{Introduction}
Road networks are spatially embedded complex networks whose link losses can
fragment access routes and alter system-wide connectivity. Exhaustive exact
evaluation becomes costly when a planner must screen many multi-link scenarios.
Conversely, a first-order perturbation is fast and interpretable; however, its local
linearity can fail under finite deletions, eigenvalue crossings, and
disconnection. This tension motivates a hybrid estimator that retains the known
spectral direction and learns only its nonlinear error.

Algebraic connectivity, the second-smallest Laplacian eigenvalue, is a global
structural quantity~\cite{fiedler1973,mohar1991}. It does not measure travel
demand, congestion, capacity, or recovery. The narrower goal here is fast
structural stress testing: given an intact weighted graph and a set of removed
edges, estimate its relative algebraic-connectivity loss. This distinction is
important because transport robustness has also been defined using system travel
time, capacity loss, accessibility, and repair trajectories
~\cite{scott2006,sullivan2010,sohouenou2021repair}.

This study asks four questions. \textbf{RQ1}: Does learning a residual over a
Fiedler approximation improve direct graph regression? \textbf{RQ2}: Does this
advantage persist under correlated disruptions and geographic domain shift?
\textbf{RQ3}: Which gains come from the explicit prior, Fiedler node feature,
coordinates, or adjacency-aware message passing? \textbf{RQ4}: Where does the
estimator fail, and how does its runtime scale?

Rather than asking whether a new graph neural network (GNN) architecture wins, the central thesis is
\emph{when a mathematically informed spectral prior helps graph learning under
distribution shift, and when it becomes a biased anchor}, meaning a systematically
miscalibrated analytical starting value that the learned correction fails to undo. The contributions
are as follows: (i) a bounded residual spectral construction for simultaneous edge
deletions; (ii) controlled feature, GCN/GraphSAGE/edge-MPNN, second-order, and reliability-context
ablations; (iii) three disruption processes on 13 cached OpenStreetMap (OSM) networks in six countries; (iv) both
synthetic-to-real zero-shot and geographically disjoint OSM transfer; (v)
area-clustered hierarchical inference, calibration, eigengap, and correction diagnostics; and
(vi) a benchmark separating eigensolver setup, amortized spectral evaluation,
and update-only cost. Importantly, the transfer experiments reveal a limitation:
the residual prior is not uniformly superior once OSM training data are
available.

\section{Related work}
\subsection{Spectral robustness of complex networks}
Fiedler introduced algebraic connectivity as a graph invariant
~\cite{fiedler1973}; later studies related Laplacian spectra to expansion,
connectivity, synchronization, and robustness
~\cite{mohar1991,chung1997,jamakovic2008}. Capacity-weighted spectral analysis
has also been used to diagnose transport-network vulnerability~\cite{bell2017}.
For a simple eigenvalue, the squared difference of the Fiedler coordinates across an
edge provides the derivative with respect to its weight~\cite{kato1995}. Such derivatives explain
infinitesimal sensitivity, whereas the present task includes several finite edge
deletions. The residual model specifically targets the missing interaction and
higher-order terms rather than replacing spectral analysis.

\subsection{Road-network disruption}
Road robustness depends on the graph representation and disruption process.
Multi-granularity analyses of empirical city networks have shown that structural
conclusions can change with representation~\cite{duan2014}. Link-based indices
using flow or capacity address functional effects~\cite{scott2006,sullivan2010},
whereas critical-scenario optimization searches for damaging link combinations
~\cite{bagloee2017}. Hazard-independent studies compare random, localized, and
targeted multi-link failures~\cite{sohouenou2021}; temporal analysis of Zurich
also demonstrates process-dependent robustness~\cite{casali2020}. Random road
graph models enable controlled topology experiments~\cite{sohouenou2020}.
Our spatial-cluster mechanism belongs to this stress-testing tradition but is a
geometric proxy, not an empirical hazard model.

Recent evidence has made scale and task definition especially important. Jana et
al.~\cite{jana2023} used an edge-based GNN to rank critical road segments after
disruption, directly motivating learning-based rapid screening. Their output is
an edge ranking, whereas ours is a graph-level spectral-loss estimate. Boeing
and Ha~\cite{boeingha2024} simulated 2.4 billion trips across more than 8,000
urban areas in 178 countries, establishing an empirical scale far beyond our
13 neighbourhood networks. Zang et al.~\cite{zang2024} predicted dynamic traffic
resilience under rainfall using a multi-granularity GNN; their dynamic, functional target is complementary to our static structural target. Reviews
distinguish connectivity, robustness, vulnerability, and recovery-based
resilience and warn against treating them as synonymous
~\cite{rivera2022,mattsson2015,zeleke2026}.

\subsection{Graph learning and analytical priors}
GCNs aggregate local neighbourhood information and are permutation equivariant
at the node level~\cite{kipf2017}. GraphSAGE supplies an inductive aggregation
backbone~\cite{hamilton2017}; GIN studies the expressive limits of neighbourhood
aggregation~\cite{xu2019}, whereas attention and message-passing frameworks expose
alternative node/edge interactions~\cite{velickovic2018,gilmer2017}. Deep Sets provides an invariant alternative that
does not propagate along edges~\cite{zaheer2017}. Graph Networks have also been
used as learnable physical simulators~\cite{sanchezgonzalez2018}, illustrating
how known structure and learned corrections can coexist. The novelty claimed
here is not bounded residual learning itself: hybrid and physics-guided
models commonly learn discrepancies around an analytical component~\cite{yuwang2024}, and the
learnable physics engine in~\cite{sanchezgonzalez2018} learns interaction
dynamics for simulation and control. The network-science contribution is more
specific. We retain the closed-form Fiedler edge derivative as an auditable
graph-level anchor, separate it experimentally from merely supplying
$u_2$ as a node feature, and test its finite multi-edge error against eigengap,
disconnection, failure geometry, and geographic shift. Correction calibration
then makes failure to override the anchor measurable rather than treating the
hybrid as an opaque accuracy device. Edge-aware modelling is especially
relevant because disruption acts on links. Jana et al.~\cite{jana2023} encoded a
road line graph for edge ranking, and Almeida et al.~\cite{almeida2025} reported
edge-aware attention for backbone-network load prediction. The present study
instead asks whether a simple node-message-passing estimator benefits from an
explicit spectral prior across domains. Recent JCN work on urban edge removal
uses diffusive transport~\cite{bowater2026}, situating algebraic connectivity
among several legitimate network-function proxies.

\section{Problem formulation and method}
Let $G=(V,E,w)$ be a connected undirected weighted graph, with combinatorial
Laplacian $L=D-A$ and eigenvalues
$0=\lambda_1\leq\lambda_2\leq\cdots$. For disrupted edges $S\subseteq E$, the
target is
\begin{equation}
y(G,S)=\operatorname{clip}\!\left(1-\frac{\lambda_2(G-S)}{\lambda_2(G)},0,1\right).
\label{eq:target}
\end{equation}
Thus, a disconnected damaged graph has $y=1$. If $u_2$ is a unit Fiedler vector
and $\lambda_2$ is simple, the edge sensitivity is
\begin{equation}
\frac{\partial\lambda_2}{\partial w_{ij}}=(u_{2,i}-u_{2,j})^2.
\end{equation}
The analytical prediction sums intact-graph derivatives:
\begin{equation}
\widehat y_{\mathrm{spec}}=\operatorname{clip}\!\left(
\frac{\sum_{(i,j)\in S}w_{ij}(u_{2,i}-u_{2,j})^2}{\lambda_2(G)},0,1\right).
\label{eq:prior}
\end{equation}

Each damaged graph supplies normalized adjacency and node features: intact
degree, failed-edge incidence, absolute normalized Fiedler coordinate, and two
normalized coordinates. Three 48-unit GCN layers feed mean--max graph pooling
and a two-layer head. The direct model predicts $y$; the residual model predicts
\begin{equation}
\widehat y_{\mathrm{res}}=\operatorname{clip}\left(
\widehat y_{\mathrm{spec}}+\tanh r_\theta(G,S),0,1\right).
\label{eq:residual}
\end{equation}
The bounded correction concentrates capacity on finite-deletion error but can
also inherit domain-specific bias from the prior.

To determine whether learning is needed beyond a higher-order analytical model, a
truncated perturbation baseline retains the 12 lowest Laplacian modes:
\begin{equation}
\Delta\lambda_2^{(2)}=u_2^\top\Delta L u_2+
\sum_{k\ne2}\frac{|u_k^\top\Delta L u_2|^2}{\lambda_2-\lambda_k}.
\end{equation}
For deletion of edges $S$, $\Delta L=-\sum_{(i,j)\in S}w_{ij}
(e_i-e_j)(e_i-e_j)^\top$. The second-order connectivity estimate is
$\lambda_2+\Delta\lambda_2^{(2)}$ and is converted to relative loss using
Eq.~\eqref{eq:target}, with the same $[0,1]$ clipping as the first-order estimate. Twelve
modes were fixed for every graph before evaluation and were not tuned on test
performance. The $k=1$ term vanishes because $\Delta L\mathbf{1}=0$;
numerically, the retained modes exclude $k=2$ from the correction sum.
We also record the relative eigengap
$\gamma=(\lambda_3-\lambda_2)/\lambda_2$, which measures proximity to the next
mode but is not supplied to any learner.

\begin{figure}[H]
\centering
\includegraphics[width=\textwidth]{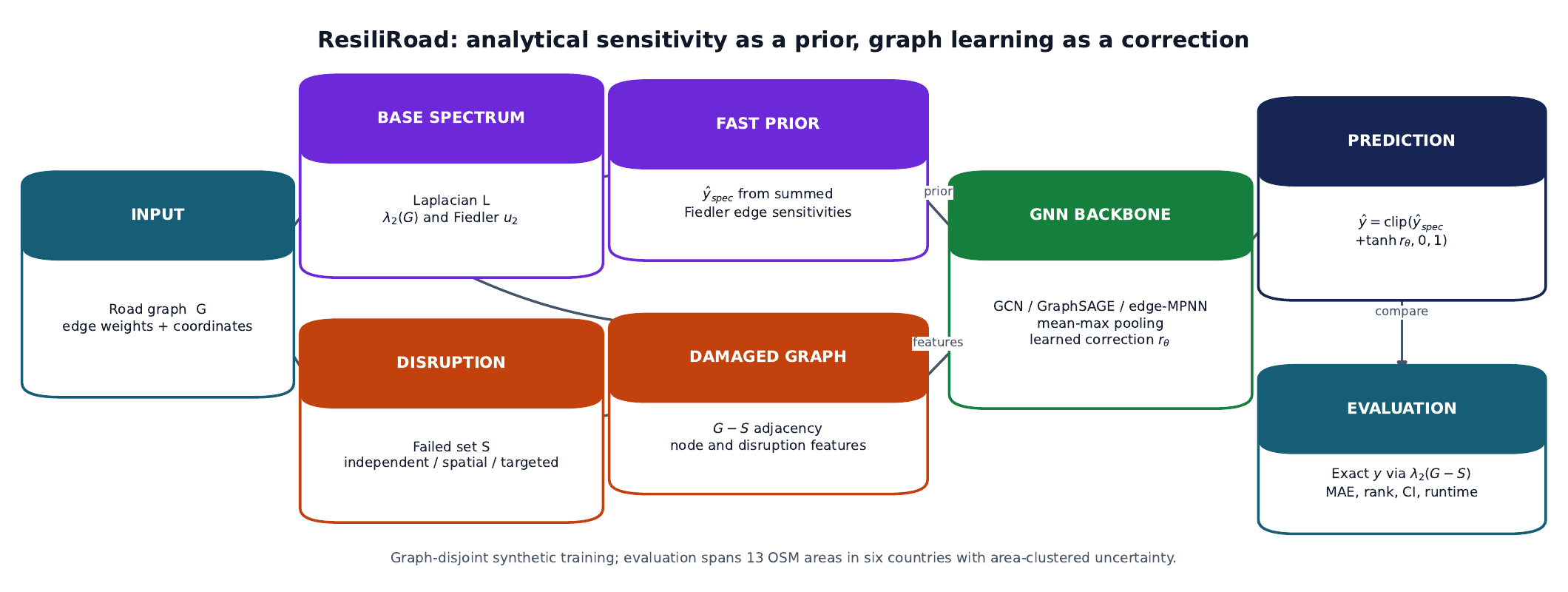}
\caption{ResiliRoad workflow. Exact damaged-graph eigendecomposition constructs
labels only. The intact Fiedler vector produces both an optional node feature
and the explicit first-order prior; the selected GNN backbone learns a bounded
correction.}
\label{fig:method}
\par\small\textit{Alt text: A flow diagram splits an intact road graph into an
exact-label branch and a fast spectral-prior branch. Damaged adjacency and node
features enter a selected GNN backbone whose residual is added to the spectral
estimate.}
\end{figure}

Figure~\ref{fig:method} summarizes the complete ResiliRoad workflow.

The ablation matrix contains direct and residual GCNs with all features, without the Fiedler feature, and without coordinates; Deep Sets and a summary-statistic MLP are
non-GCN baselines. A further reliability-context residual appends the prior,
normalized failure count, density, and graph size to the prediction head. It
does not use damaged connectivity or any label-derived flag. As a stronger
architecture control, GraphSAGE concatenates each node state with a normalized
neighbourhood aggregate before each of the three learned transformations. Its direct
and residual variants use the same features, pooling, optimizer, split, and
stopping rule as the GCN.
The edge-aware MPNN instead forms
$m_{ij}=\phi(h_i,h_j,e_{ij})$ over every intact candidate link, where
$e_{ij}$ contains normalized conductance, inverse-conductance length, a failed
edge indicator, and normalized $(u_{2,i}-u_{2,j})^2$. Mean incoming messages
update node states in three layers. Keeping failed links as candidates allows
the model to observe deletions explicitly rather than only through damaged
adjacency and node incidence.

\section{Experimental design}
\subsection{Synthetic graphs and disruption processes}
Connected random geometric graphs contained 35--65 nodes and used inverse-length edge
conductances. The original two-mode experiment generated 800 scenarios per mode
and seed, and the expanded three-mode experiment generated 400 per mode and seed.
All base-graph families were assigned to training, validation, or testing. In each
scenario, one to eight edges were removed. Independent failures sampled edges uniformly.
For spatial-cluster failures, a random epicentre was chosen and the required
number of edges with the nearest geometric midpoints was removed. Both modes used the same
failure-count distribution.
The targeted regime sampled without replacement with probability proportional to
unweighted edge-betweenness centrality. This procedure targeted globally important links
without using the Fiedler sensitivity that defines the analytical prior.

\subsection{OpenStreetMap networks}
OSMnx~\cite{boeing2017} was used to obtain drivable networks around 13 areas in
Vietnam, Singapore, Malaysia, Thailand, Taiwan, and Japan. Five 650-m
neighbourhoods preserve the original evaluation; eight 1.2--1.6-km areas add
larger and morphologically diverse networks. Directed multigraphs were simplified, converted to undirected
simple graphs, and restricted to the largest connected component. Conductance
is $100/\mathrm{length}_{\mathrm m}$. Cached GraphML freezes the exact instances
because live OSM data evolve.

\begin{figure}[H]
\centering
\includegraphics[width=\textwidth]{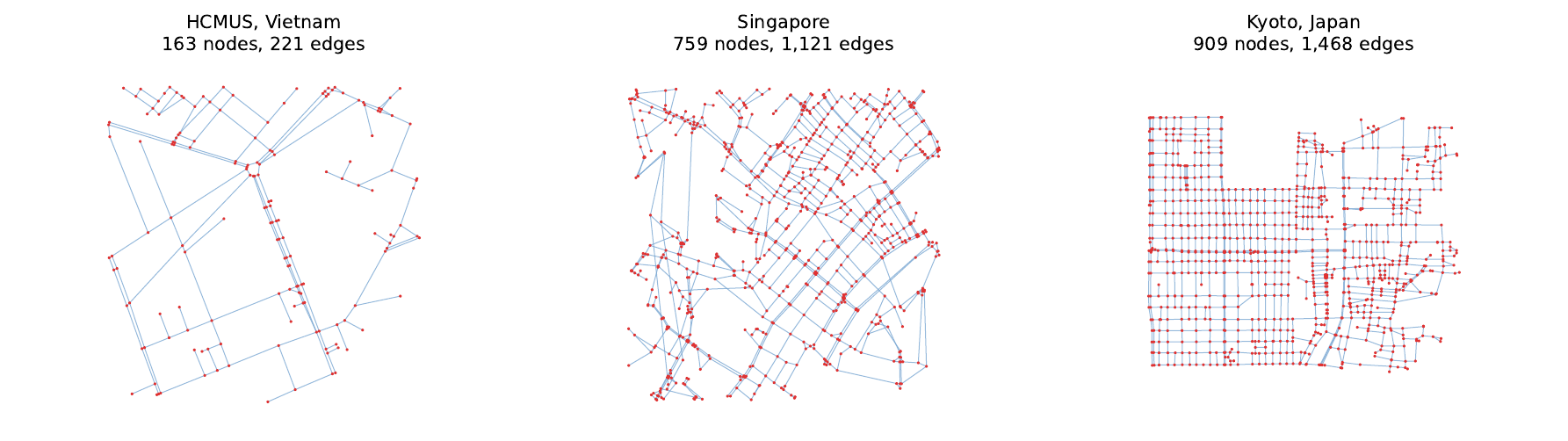}
\caption{Three processed OSM base networks illustrating contrasting scale and
street morphology: HCMUS, Singapore, and Kyoto. Disruption scenarios remove
subsets of the displayed links. Map data copyright OpenStreetMap contributors.}
\label{fig:osm}
\par\small\textit{Alt text: Three road graphs appear in one row with red
intersection nodes and pale-blue links. HCMUS is small and branching,
Singapore is dense and irregular, and Kyoto has a visibly stronger grid.}
\end{figure}

Table~\ref{tab:osm} summarizes the OSM networks, while
Figure~\ref{fig:osm} illustrates three representative street morphologies.

\begin{table}[H]
\centering
\caption{OpenStreetMap networks used in the study.}
\label{tab:osm}
\resizebox{\textwidth}{!}{\begin{tabular}{llrrrr}
\toprule Area & Country & Radius (m)&Nodes & Edges & Density\\ \midrule
HCMUS&Vietnam&650&163&221&.0167\\ VIASM&Vietnam&650&206&301&.0143\\
Da Nang&Vietnam&650&142&203&.0203\\ Can Tho&Vietnam&650&114&178&.0276\\
Da Lat&Vietnam&650&48&55&.0488\\ Singapore&Singapore&1,400&759&1,121&.00390\\
Kuala Lumpur&Malaysia&1,500&730&1,059&.00398\\ George Town&Malaysia&1,600&936&1,334&.00305\\
Bangkok&Thailand&1,400&1,072&1,457&.00254\\ Chiang Mai&Thailand&1,600&1,259&1,661&.00210\\
Taipei&Taiwan&1,300&965&1,546&.00332\\ Kyoto&Japan&1,500&909&1,468&.00356\\
Tokyo&Japan&1,200&1,108&1,675&.00273\\ \bottomrule
\end{tabular}}
\end{table}

Two transfer protocols answer different questions. In \emph{zero-shot OSM}, all
learned models train only on synthetic graphs and test on 100 scenarios per
site, mode, and seed. In \emph{leave-one-area-out OSM}, one site is test, the
next site in a fixed rotation is validation, and the remaining three sites are
training. Forty scenarios per site and mode are generated for each of five
seeds; the direct and residual GCNs train for at most 25 epochs. No base area is
shared across train, validation, and test in a fold.
An expanded zero-shot protocol trains GCN, GraphSAGE, and edge-MPNN direct and
residual variants on 400 synthetic scenarios per failure mode and evaluates 20
scenarios per area, mode, and seed across all 13 OSM areas. The original and
expanded protocols are reported separately; the latter supplies geographic
rather than merely random-seed replication.

The final OSM-to-OSM experiment uses leave-one-country-out (LOCO) transfer. In
each of six folds, every area from one country is held out for testing, the next
country in the manifest's fixed cyclic order is reserved for validation, and
the other four countries train the models. In the primary matched protocol, a
separate direct and residual GCN is trained for each failure mode for at most 12
epochs. A joint three-mode version is retained as a sensitivity analysis. Each
seed supplies eight scenarios per area and mode. Country is the outer inferential unit and
seed is nested within country; neither areas within a held-out country nor
scenarios are treated as independent country replications.
The same \texttt{perf\_counter} timing around optimization gave summed per-fit
wall times of 6.49 h for 60 joint fits and 0.49 h for 180 matched fits, excluding
label generation and evaluation. Because these were separate executions rather
than a controlled timing benchmark, the unadjusted totals document provenance
and are not evidence of comparative efficiency.

Two additional robustness controls delimit competing explanations. First, we
replay every expanded OSM disruption and compute two topology-only transport
proxies: relative loss of the largest connected component and relative loss of
mean inverse shortest-path length over 64 fixed origin--destination pairs per
area. The latter uses inverse conductance as link length. These are not substitutes
for demand- and capacity-based robustness indices~\cite{scott2006,sullivan2010};
they test whether common connectivity and accessibility summaries approximate
the stated $\lambda_2$-loss target. Second, a size-only control uses the same
connected planar lattice family, edge-weight rule, 1--8-edge failure policy,
and three failure regimes at 50, 100, 200, 400, 800, and 1,200 nodes. For each
size and regime it evaluates 60 scenarios over five seeds and two graph replicates.

\subsection{Statistics, calibration, and runtime}
For synthetic graph populations, seeds 11, 22, 33, 44, and 55 define independent
generator replicates. For OSM, the road area is the outer sampling cluster and
seed is nested within area. Hierarchical intervals were obtained by resampling
areas with replacement and then one seed realization within each sampled area
10,000 times. An effect has inferential support here only when its area- or
country-clustered 95\% interval excludes zero; otherwise its sign is reported
strictly as a descriptive point-estimate trend. This convention is especially
important for six-cluster LOCO and is not repaired by extra scenario or seed
rows. Paired comparisons preserve method pairing within area and seed. The
five-area leave-one-area-out analysis is re-estimated with the same area-outer
hierarchy. No scenario-level $p$-value is used.
Continuous calibration is summarized by regressing target on prediction within
each seed, reporting slope, intercept, and signed mean error. Error slices by
connectivity, failure count, density/site, and spectral clipping are diagnostic.

The original OSM benchmark uses dense symmetric eigendecomposition and
one-scenario neural inference on CPU. A separate controlled benchmark uses sparse symmetric eigensolver computations on random geometric graphs with 50--800 nodes. It
reports exact damaged-graph solution, update-only Fiedler sensitivity, and
amortized spectral cost (one intact setup spread over ten scenarios). These
denser graphs are computational stress tests rather than road-realistic samples.

\section{Results}
\subsection{Graph-disjoint synthetic and zero-shot OSM results}
Table~\ref{tab:primary} summarizes the graph-disjoint synthetic and zero-shot
OSM results.
\begin{table}[H]
\centering\small
\caption{MAE as seed mean [95\% seed-bootstrap interval].}
\label{tab:primary}
\resizebox{\textwidth}{!}{%
\begin{tabular}{lcc|cc}
\toprule &\multicolumn{2}{c|}{Synthetic}&\multicolumn{2}{c}{Zero-shot OSM}\\
Method&Independent&Spatial&Independent&Spatial\\ \midrule
Spectral&.069 [.048,.095]&.144 [.102,.183]&.502 [.482,.521]&.650 [.643,.657]\\
Summary MLP&.065 [.051,.079]&.108 [.082,.135]&.281 [.240,.320]&.347 [.269,.425]\\
Deep Sets&.068 [.058,.078]&.105 [.085,.126]&.114 [.105,.123]&.143 [.122,.164]\\
Direct GCN&.077 [.055,.098]&.081 [.067,.097]&.106 [.100,.112]&.102 [.077,.148]\\
Residual GCN&\textbf{.052 [.036,.070]}&\textbf{.065 [.049,.079]}&
\textbf{.097 [.091,.102]}&\textbf{.090 [.071,.124]}\\ \bottomrule
\end{tabular}}
\end{table}

The residual model improves direct-GCN MAE by paired differences 0.0253
[0.0106, 0.0400] and 0.0160 [0.0048, 0.0241] on synthetic independent and
spatial failures. The improvements remain positive in zero-shot OSM: 0.0085
[0.0039, 0.0131] and 0.0121 [0.0026, 0.0247]. Yet the analytical prior alone
has poor OSM calibration despite useful rank information. Removing Fiedler or
coordinates from residual-full yields intervals containing zero; individual
node-feature necessity is therefore not established.

\subsection{Geographically blocked transfer}
Table~\ref{tab:loao} reports the geographically blocked leave-one-area-out
results.
\begin{table}[H]
\centering
\caption{Leave-one-area-out OSM performance, area mean [95\% hierarchical
area-outer interval]. Each fold uses three training, one validation, and one test area.}
\label{tab:loao}
\begin{tabular}{lcc}
\toprule Method&Independent MAE&Spatial-cluster MAE\\ \midrule
Direct GCN&\textbf{.135 [.089,.199]}&\textbf{.109 [.067,.158]}\\
Residual GCN&.161 [.089,.319]&.144 [.059,.352]\\ \bottomrule
\end{tabular}
\end{table}

The point-estimate ranking reverses when training is moved into the OSM domain. Direct GCN has
mean $R^2$ 0.714 and 0.747, compared with 0.550 and 0.529 for the residual
model. The paired residual-minus-direct MAE differences are 0.0261
[-0.0357, 0.1572] and 0.0350 [-0.0479, 0.2320]; both area-outer intervals include zero,
so five areas do not establish direct-model superiority. Residual performance is seed-sensitive: independent-failure MAE ranges
from 0.092 to 0.265. Calibration partly explains this instability. Direct-GCN
slopes stay between 0.91 and 1.06; residual slopes fall to 0.58 and 0.69 in the
two poorest seeds, with substantial underprediction. Thus the spectral prior is
helpful under one training distribution but can become a shortcut or a biased
anchor under another.

\subsection{Expanded leave-one-country-out transfer}
The matched six-country LOCO experiment does not reproduce a universal
OSM-to-OSM reversal (Table~\ref{tab:loco}). Residual GCN improves point-estimate
MAE for independent and targeted failures, whereas direct GCN is better for
spatial clusters. Only the targeted interval excludes zero. The joint-training
sensitivity analysis has the same three signs (differences .0124, $-.0335$,
and .0271), so this pattern is not explained by multi-regime training. Thus the
five-area reversal is evidence of instability, not evidence that direct
prediction generally wins after training on real roads. Across-country transfer
is failure-regime and country dependent, and six outer clusters still limit
precision.

\begin{table}[H]
\centering\small
\caption{Matched per-regime leave-one-country-out OSM transfer. One model is
trained per failure regime. MAE values average countries and seeds; differences
are direct minus residual with 95\% country-outer, seed-within-country
hierarchical intervals.}
\label{tab:loco}
\resizebox{\textwidth}{!}{%
\begin{tabular}{lrrr}
\toprule Failure regime&Direct GCN&Residual GCN&Difference [95\% interval]\\ \midrule
Independent&.182&.145&.0372 [-.1475,.1679]\\
Spatial cluster&.061&.149&-.0874 [-.2942,.0292]\\
Targeted betweenness&.258&.196&\textbf{.0622 [.0169,.1153]}\\ \bottomrule
\end{tabular}}
\end{table}

\begin{figure}[H]
\centering
\includegraphics[width=.86\textwidth]{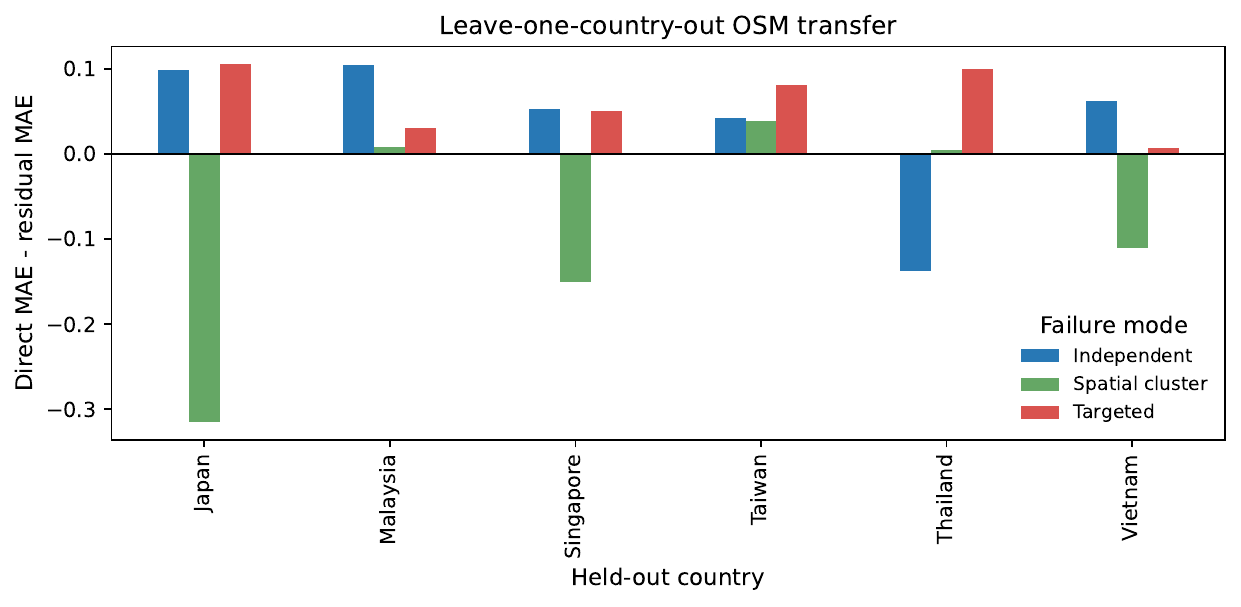}
\caption{Matched per-regime direct-minus-residual GCN MAE for each held-out
country, averaged over five seeds. Positive values favour the residual model.}
\label{fig:loco}
\par\small\textit{Alt text: A grouped bar chart shows residual gains varying
substantially by held-out country and failure regime, with both positive and
negative bars.}
\end{figure}

Figure~\ref{fig:loco} shows the corresponding country-level
direct-minus-residual differences across failure regimes.

\subsection{Reliability-context negative ablation}
Adding graph size, density, failure count, and spectral estimate to the residual
head reduces synthetic spatial MAE by 0.0193 [0.0120, 0.0266] and has an
inconclusive $-0.0034$ [-0.0159, 0.0111] change for synthetic independent
failure (negative means improvement). It increases zero-shot OSM spatial MAE by
0.0304 [0.0067, 0.0579]; the independent increase is 0.0452
[-0.0029, 0.1019]. Simple reliability metadata therefore encourages
distribution-specific calibration rather than robust transfer.

\subsection{Backbone robustness}
Replacing GCN propagation with GraphSAGE tests whether the main conclusion is
an artefact of a weak backbone. Appendix Table~\ref{tab:backbone} shows that this is not
the case: residual learning improves point-estimate MAE for both backbones in
all four zero-shot conditions. Paired direct-minus-residual differences for
GraphSAGE are 0.0197 [0.0071, 0.0332] and 0.0190 [0.0107, 0.0273] on synthetic
independent and spatial failures, and 0.0069 [0.0032, 0.0096] and 0.0239
[-0.0073, 0.0697] on OSM. The final OSM-spatial interval crosses zero, so the
effect is not declared established there.

\subsection{Cross-country, targeted, and edge-aware evaluation}
Table~\ref{tab:expanded} reports the expanded 13-area experiment. Positive
paired differences favour the residual formulation. Residual point estimates
improve all nine backbone--failure combinations, but area-level uncertainty is
material: intervals exclude zero only for GCN under spatial failure and
GraphSAGE under independent and targeted failure. The edge-aware MPNN does not
outperform the node-message-passing models; for example, its residual MAE is
.163, .199, and .156, compared with residual-GCN MAE .085, .057, and .135.
Thus explicit failed-edge attributes are an architecture control, not an
automatic accuracy gain.

\begin{table}[H]
\centering\small
\caption{Expanded zero-shot OSM paired improvement (direct minus residual
MAE), area mean [95\% hierarchical interval].}
\label{tab:expanded}
\resizebox{\textwidth}{!}{\begin{tabular}{lccc}
\toprule Backbone&Independent&Spatial cluster&Targeted betweenness\\ \midrule
GCN&.0167 [-.0019,.0338]&\textbf{.0391 [.0151,.0662]}&.0097 [-.0095,.0268]\\
GraphSAGE&\textbf{.0253 [.0114,.0425]}&.0159 [-.0043,.0365]&\textbf{.0257 [.0095,.0446]}\\
Edge-MPNN&.0042 [-.1210,.1227]&.0128 [-.1489,.1575]&.0359 [-.0412,.1113]\\ \bottomrule
\end{tabular}}
\end{table}

\subsection{Transport-proxy and graph-size controls}
The size-only experiment is a within-family sensitivity analysis, not a
morphology-matched synthetic-to-OSM transfer test.
The transport-topology proxies do not approximate the spectral target closely.
Area-mean MAE for OD-efficiency loss is .396, .582, and .244 under independent,
spatial, and targeted failure; largest-component loss gives .403, .592, and
.256. Area-clustered intervals are shown in Appendix Figure~\ref{fig:revisioncontrols}.
These controls show that the learned gains are not obtained by relabelling a
component-size or shortest-path summary. They do not show that algebraic
connectivity is a better measure of traffic service, because no demand,
capacity, or assignment data enter the experiment.

Within the fixed planar-lattice family, first-order spectral MAE decreases from
.115 to .002 for independent failures and from .091 to .002 for targeted
failures between 50 and 1,200 nodes. Spatial-cluster MAE remains non-monotone
(.478 at 50 nodes and .300 at 1,200) because localized removal often disconnects
the lattice. Under the fixed 1--8-edge policy, increasing size alone therefore
does not reproduce the broad real-network prior miscalibration. This control
does not fully identify morphology, but size is no longer the sole untested
explanation.

\subsection{Eigengap, analytical order, and learned correction}
The truncated second-order perturbation baseline improves over the first-order baseline by
.00284 [.00050,.00673], .00359 [.00081,.00756], and .00533
[.00133,.01187] MAE under independent, spatial, and targeted disruption,
respectively. These gains are consistent but too small to close the gap to the
best learned estimators, answering why a learned correction remains useful.

The relative eigengap $\gamma=(\lambda_3-\lambda_2)/\lambda_2$ is informative
but not a sufficient gate. First-order OSM MAE is .444, .369, and .422 in low,
middle, and high graph-level eigengap tertiles: near-degeneracy is harder than
the middle tertile, but the relationship is non-monotone. Residual improvement
also does not vanish uniformly at small gaps.

Mechanistic correction diagnostics support a more specific biased-anchor
interpretation. Under targeted transfer, regression slopes relating the learned correction to the needed correction are .512 (GCN), .580 (GraphSAGE), and .486 (edge-MPNN), compared with .843, .834, and .617 under independent failure. The residual
therefore tends to shrink the correction precisely in the targeted regime,
rather than fully undoing a miscalibrated analytical anchor.

\begin{figure}[H]
\centering
\includegraphics[width=.96\textwidth]{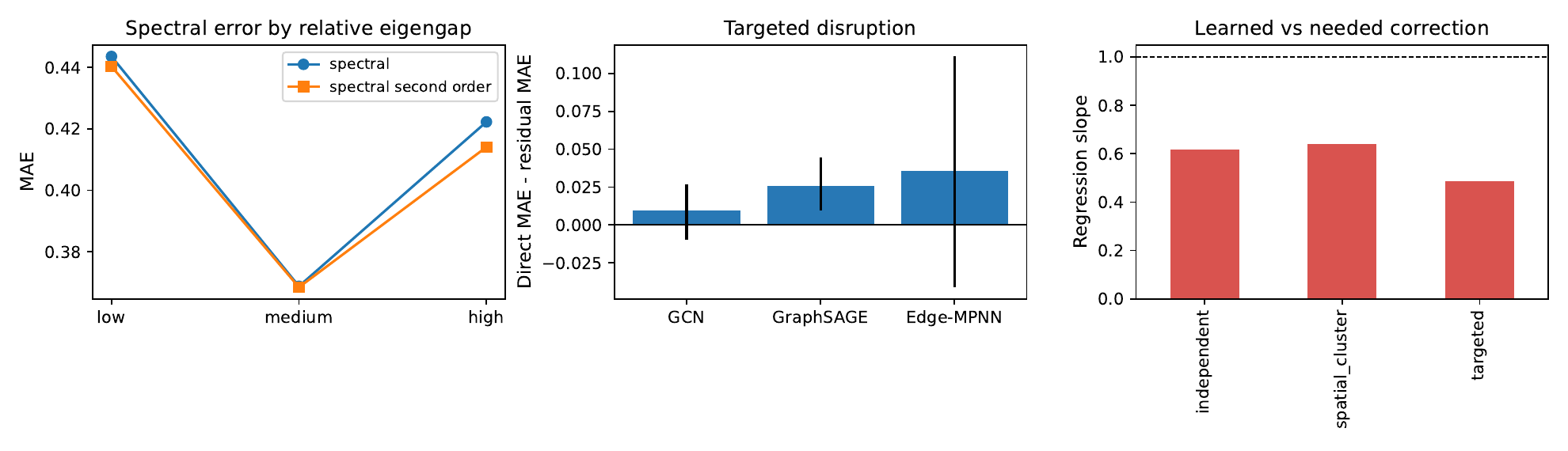}
\caption{Expanded diagnostic results. (a) First- and second-order spectral
error across relative-eigengap tertiles. (b) Area-level direct-minus-residual
MAE under targeted disruption with hierarchical intervals. (c) Learned versus
needed residual correction.}
\label{fig:jcn2}
\par\small\textit{Alt text: Three panels compare residual improvements across
backbones and disruptions, spectral errors across eigengap groups, and the
calibration of learned corrections against ideal corrections.}
\end{figure}

Figure~\ref{fig:jcn2} summarizes the eigengap, targeted-disruption, and
correction-calibration diagnostics.

\begin{figure}[H]
\centering
\includegraphics[width=.94\textwidth]{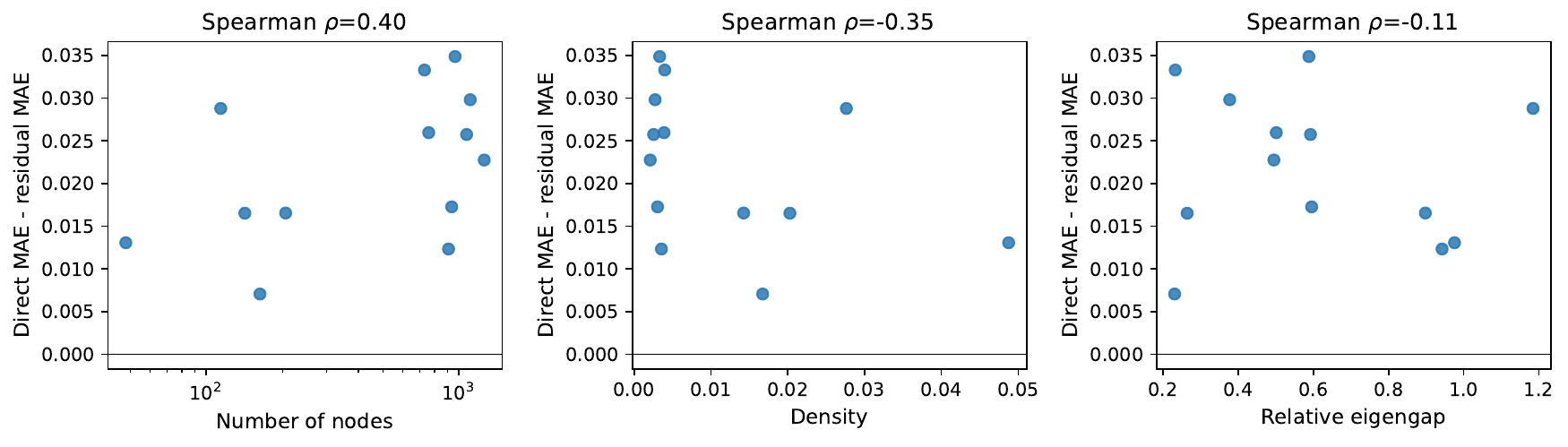}
\caption{Descriptive residual-GCN gain versus OSM graph size, density, and
relative eigengap. Each point is one area averaged over seeds and failure
regimes; positive values favour residual prediction.}
\label{fig:structuregain}
\par\small\textit{Alt text: Three scatter plots show residual gain against
node count, density, and eigengap. Points are dispersed without a strong
monotone relationship, especially for eigengap.}
\end{figure}

A descriptive area-level check finds Spearman correlations of residual-GCN
gain with node count, density, and relative eigengap of 0.40, $-0.35$, and
$-0.11$, respectively (Figure~\ref{fig:structuregain}). With only 13 purposively
selected areas these are not inferential or causal estimates, but they show no
simple collapse of the residual gain as graph size increases and no single
structural variable explains it.

\subsection{Runtime and error regimes}
On the five small OSM networks, dense exact recomputation, first-order update,
and residual inference average 5.094, 0.018, and 0.674 ms per scenario on CPU.
In the 50--800-node sparse control, exact time rises from 8.37 to 92.61 ms,
whereas residual-GCN inference rises from 0.67 to 5.92 ms; intermediate values
are archived with the code.

To test the actual screening motivation, a second sparse benchmark uses
connected planar road-like graphs with 1,000, 2,000, 5,000, 10,000, and 20,000
nodes, with $|E|/|V|\approx2.1$ (mean degree $\bar d\approx4.2$). The connected
near-square grid receives planar diagonals with probability 0.12 and
inverse-length conductance; each disruption removes eight random edges. At
20,000 nodes, sparse exact recomputation
requires 1,302.4 ms per damaged graph. Intact eigensolver setup costs 1,335.3
ms once; amortized over 1,000 screened scenarios, the prior costs 1.41 ms,
sparse GNN inference 21.76 ms, and the combined path 23.17 ms. Thus the measured
screening path is 56 times faster at the largest size, conditional on reusing
one intact graph. Only three scenarios on each of two graphs are timed per size,
and none is used for accuracy training; this is a scaling check, not evidence
of predictive generalization.

\begin{figure}[H]
\centering
\includegraphics[width=.92\textwidth]{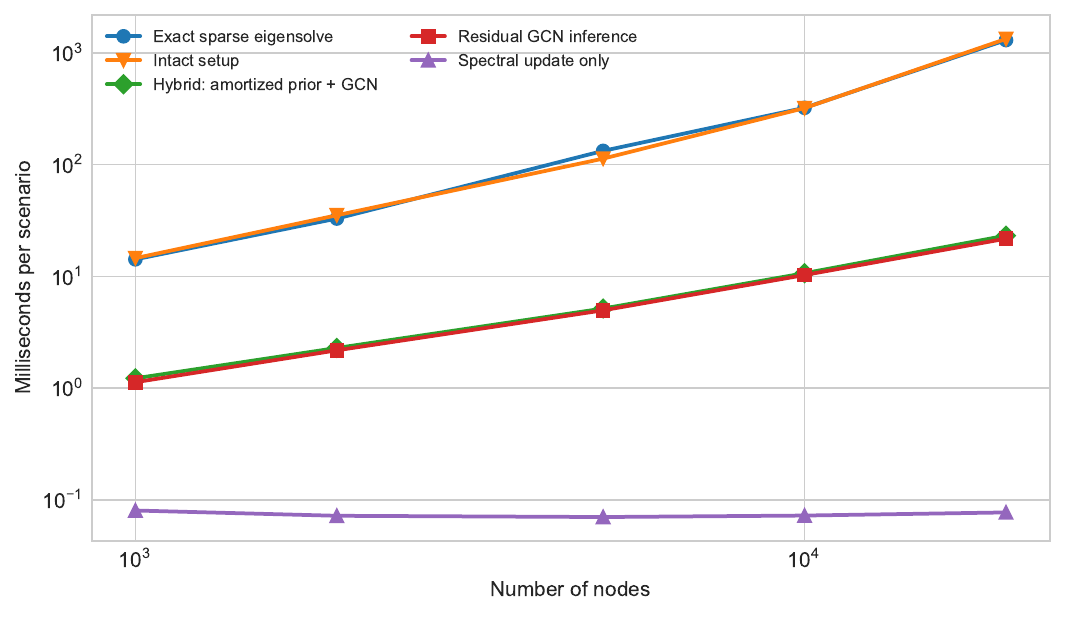}
\caption{Sparse CPU scaling from 1,000 to 20,000 nodes. Exact recomputation is
timed per damaged graph; the hybrid curve combines intact spectral setup
amortized over 1,000 scenarios with first-order evaluation and sparse GNN
inference. Points are means over two graphs and three scenarios per graph.}
\label{fig:scaling}
\par\small\textit{Alt text: A log--log runtime plot shows exact sparse
eigendecomposition rising to about 1.3 seconds at 20,000 nodes, while the
amortized hybrid path rises to about 23 milliseconds.}
\end{figure}

The corresponding scaling curves are shown in Figure~\ref{fig:scaling}.

In zero-shot OSM, residual-full MAE is 0.070 for connected damaged graphs and
0.112 after disconnection. Site errors range from 0.041 (Can Tho) to 0.189
(Da Lat). The densest site is also the smallest and hardest, so density cannot
be interpreted causally. No evaluated first-order estimate clips at one; the
requested clipped/unclipped comparison is therefore undefined. The largest
leave-one-area-out errors are retained in a machine-readable table instead of
being removed as outliers.

\section{Discussion}
Across the 12 primary paired OSM comparisons (nine expanded zero-shot and three
matched LOCO), four have inferential support because their area- or
country-clustered 95\% intervals exclude zero; the other eight are treated only
as descriptive trends. This convention governs the interpretation below.

The central result is a condition rather than a new architectural claim. A spectral
residual improves point estimates for GCN, GraphSAGE, and edge-MPNN on the
expanded zero-shot OSM test. It combines the prior's ranking
signal with learned nonlinear calibration. However, blocked OSM transfer and
reliability-context ablation show that the same inductive bias can amplify
domain-specific error. The prior can then behave as a shortcut or biased anchor:
optimization remains close to a convenient analytical signal even when its
calibration has shifted. Yet matched LOCO results are mixed rather than a
consistent direct-model victory: the residual helps two failure regimes and
hurts one, with country-level support only for targeted failures. The same
sign pattern under joint training rules out training-regime pooling as its
cause. Consequently, the evidence documents domain- and
regime-sensitive behaviour in this sample, not a general causal effect of
training domain or an architectural guarantee.

The study also separates the explicit prior from Fiedler node information.
Neither Fiedler nor coordinate removal produces a stable loss, whereas replacing
the residual architecture does. The robust ingredient in the first protocol is
therefore how mathematical and learned estimates are combined, not merely the
presence of an eigenvector channel. Targeted-failure correction slopes show
that a biased anchor is a measurable calibration failure, while the
non-monotone eigengap result warns against using eigengap alone as a trust
rule. The poor Deep Sets ranking in several
regimes supports adjacency-aware propagation, while the summary MLP shows that
coarse graph statistics are insufficient.

For practical deployment, the model should be used as a screening layer rather
than as a replacement for exact analysis or traffic simulation. An operator can
precompute the intact-network spectrum, rank large scenario batches with the
hybrid estimator, and recompute exact metrics for the highest-risk or most
uncertain cases. Because calibration changes across geography and failure
regime, deployment should include blocked validation on representative local
areas and should fall back to exact computation when that validation fails.

\subsection{Limitations}
Thirteen purposively selected neighbourhoods in six Asian countries improve
geographic and morphological coverage but do not establish global or
city-scale predictive generality; they span only 48--1,259 nodes. Their size,
density, country, and morphology remain partly confounded. The
synthetic geometric generator and large-scale graphs are controlled proxies,
not metropolitan OSM accuracy datasets. Spatial clusters are not calibrated to flood,
landslide, earthquake, construction, or conflict data. Algebraic connectivity
is structural: it omits direction, demand, congestion, capacity, travel time,
accessibility, and restoration. Only five seed realizations per area limit
within-area precision, while 13 areas limit geographic inference. The
hierarchical zero-shot analysis treats area as the outer unit, while LOCO uses
country; neither can compensate for purposive selection. LOCO has only six
country clusters and uses one fixed cyclic validation assignment, so its wide
intervals and sensitivity to validation geography must be retained. No
probabilistic predictive intervals are
produced. Sparse batched neural
implementations and spectral sparsification~\cite{spielman2011} may change
runtime rankings at still larger scales.

Future work should pre-register larger metropolitan regions, use nested blocked
validation across countries, and disentangle graph size from density. Hazard rasters
could define correlated failures, while traffic assignment could test whether
structural loss predicts service degradation. A mixture-of-experts model or learned
gating may decide when to trust the prior; conformal methods could add
graph-population prediction regions. These extensions should preserve area-level
replication rather than inflate sample size with correlated scenarios.

\section{Conclusion}
Residual spectral graph learning offers an interpretable, fast estimator of
finite multi-edge connectivity loss whose observed advantage varies with
training domain and failure regime. The method shows heterogeneous point-estimate gains across
GCN, GraphSAGE, and edge-MPNN in the expanded zero-shot study. OSM-to-OSM
results vary by protocol and failure regime: the
five-area reversal does not persist uniformly in matched six-country LOCO
transfer, and joint training yields the same sign pattern.
Reporting both outcomes, along with ablations, hierarchical
uncertainty, targeted attacks, eigengap and second-order controls, correction
calibration, error cases, and separated runtime costs, provides a
reproducible account of when a Fiedler prior helps and when it does not.

\section*{Data and code availability}
Code, cached OSM-derived GraphML, manifests, experiment scripts, environment
versions, compact result tables, and release metadata are available as
ResiliRoad version 1.0.0, permanently archived on Zenodo~\cite{resiliroad}.
The OSM data are licensed under the Open Data Commons Open Database License and are
attributed to OpenStreetMap contributors~\cite{osm}. The immutable archived
release fixes the code, data, configurations, and compact outputs used for the
submitted results; ongoing development is hosted at
\url{https://github.com/lee-vtruong/ResiliRoad}.

\section*{Funding}
This research received no external funding.

\section*{Conflict of interest}
The author declares no conflict of interest.

\section*{Ethics approval}
Ethics approval was not required because this study did not involve human
participants, human data, animals, or personally identifiable information.

\section*{Acknowledgements and AI-use disclosure}
Generative AI tools were used for language editing and software-development
assistance. The author designed the study, verified the implementation and
results, and takes responsibility for the manuscript.

\appendix
\section{Reproducibility details}
All models use AdamW, mean-squared error, validation early stopping, and one
PyTorch compute thread. Core runs use at most 80 epochs; area-blocked OSM
transfer uses at most 25 epochs and LOCO uses at most 12 epochs. Software versions are Python 3.13.11, NumPy 2.4.4, SciPy
1.17.1, NetworkX 3.6.1, pandas 2.2.3, scikit-learn 1.9.0, PyTorch 2.11.0
(CPU), Matplotlib 3.11.1, OSMnx 2.1.1, and GeoPandas 1.1.4. The machine had 16
logical AMD64 CPUs; CUDA was unavailable. Scripts record exact seeds and emit
scenario predictions, training histories, fold assignments, runtime summaries,
bootstrap tables, calibration diagnostics, and worst cases.
The architecture-control runs use identical data and training settings for GCN
and GraphSAGE. Expanded runs add an edge-MPNN whose directed messages receive
normalized conductance, inverse-conductance length, failed-edge status, and
normalized squared Fiedler difference. Targeted scenarios sample without
replacement in proportion to intact-graph edge betweenness. For each OSM area,
20 scenarios per mode and seed are evaluated; hierarchical resampling uses area
as the outer cluster and seed within area. Second-order perturbation uses the
lowest 12 Laplacian modes, and eigengap tertiles are assigned at graph level.
Large-scale runtime graphs use a normalized near-square lattice,
four-neighbour grid backbone, independent northwest diagonals with probability
0.12, and inverse-length conductance; the deterministic grid guarantees a
connected base graph and yields mean degree approximately 4.2. The sparse
\texttt{eigsh} solver uses a fixed starting vector. The intact setup cost is
reported both separately and amortized over 1,000 scenarios, preventing
preprocessing from being hidden in the speed comparison.

The transport control replays the stored OSM scenario seeds and checks failure
counts against the original predictions before computing largest-component and
64-pair inverse-shortest-path losses. The size control uses two planar-lattice
graphs per size, six scenarios per regime and seed, and five seeds. Machine-readable
scenario predictions and summaries accompany the plotting script.

\begin{table}[H]
\centering\small
\caption{Architecture control: MAE, seed mean [95\% seed-bootstrap interval].}
\label{tab:backbone}
\resizebox{\textwidth}{!}{\begin{tabular}{llcccc}
\toprule Domain&Failure&Direct GCN&Residual GCN&Direct GraphSAGE&Residual GraphSAGE\\ \midrule
Synthetic&Independent&.077 [.054,.096]&.052 [.036,.071]&.066 [.050,.081]&\textbf{.046 [.034,.063]}\\
Synthetic&Spatial&.081 [.067,.097]&\textbf{.065 [.049,.079]}&.084 [.068,.101]&.065 [.047,.085]\\
OSM&Independent&.106 [.100,.111]&\textbf{.097 [.091,.102]}&.104 [.098,.108]&.097 [.091,.103]\\
OSM&Spatial&.102 [.077,.148]&.090 [.071,.124]&.113 [.082,.164]&\textbf{.089 [.079,.099]}\\ \bottomrule
\end{tabular}}
\end{table}

\begin{figure}[H]
\centering
\includegraphics[width=.96\textwidth]{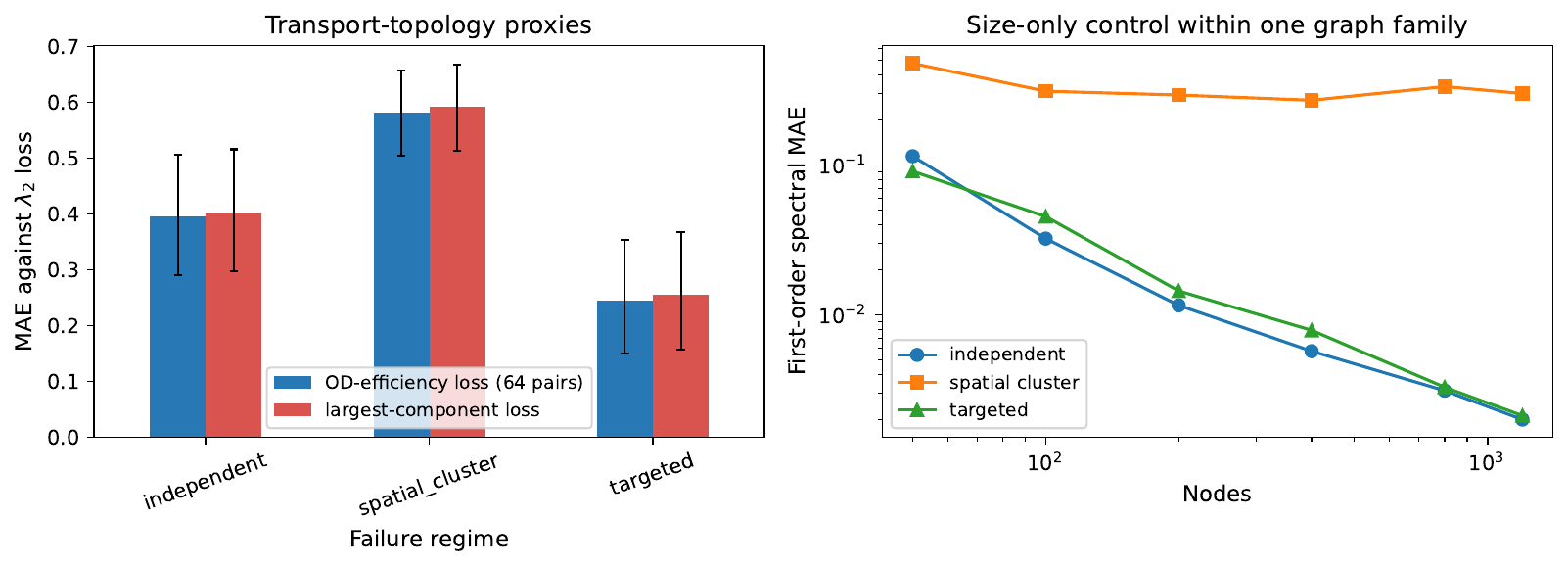}
\caption{Additional robustness controls. (a) Topology-only transport-proxy error
against the algebraic-connectivity-loss target. (b) First-order spectral error
as graph size changes within one planar road-like graph family.}
\label{fig:revisioncontrols}
\par\small\textit{Alt text: The left panel shows high proxy errors across three
failure regimes. The right log--log panel shows spectral error decreasing with
size for independent and targeted failures but remaining high and non-monotone
for spatial failures.}
\end{figure}

Table~\ref{tab:modelbudget} summarizes the parameter counts and common training
settings across the backbones.

\begin{table}[H]
\centering\small
\caption{Fairness controls for the expanded architecture comparison. Parameter
counts are identical for direct and residual variants of each backbone.}
\label{tab:modelbudget}
\resizebox{\textwidth}{!}{%
\begin{tabular}{lrrrrrr}
\toprule Backbone&Parameters&Hidden&Layers&Learning rate&Max epochs&Patience\\ \midrule
GCN&9,745&48&3&.002&40&15\\
GraphSAGE&14,593&48&3&.002&40&15\\
Edge-MPNN&40,609&48&3&.002&40&15\\ \bottomrule
\end{tabular}}
\end{table}
All three use the same scenarios, splits, AdamW optimizer, MSE objective,
weight decay $10^{-4}$, validation rule, and random seeds. Hyperparameters were
fixed before the expanded test and were not selected separately to favour a
backbone. Edge-MPNN has substantially more parameters, so its weaker result
cannot be attributed to an intentionally smaller capacity; it may nevertheless
benefit from architecture-specific tuning, which was outside the scope of this control.

\end{document}